\documentclass[conference]{IEEEtran}

\usepackage{cite}
\usepackage{amsmath,amssymb,amsfonts}
\usepackage{algorithmic}
\usepackage{graphicx}
\usepackage{textcomp}
\usepackage{xcolor}
\usepackage{caption}
\usepackage{subcaption}
\usepackage{comment}
\usepackage{tikz}
\PassOptionsToPackage{hyphens}{url}\usepackage{hyperref}

\def\BibTeX{{\rm B\kern-.05em{\sc i\kern-.025em b}\kern-.08em
    T\kern-.1667em\lower.7ex\hbox{E}\kern-.125emX}}

\newcommand\copyrighttext{%
	\footnotesize \textcopyright 2025 IEEE. Personal use of this material is permitted. Permission from IEEE must be obtained for all other uses, in any current or future media, including reprinting/republishing this material for advertising or promotional purposes, creating new collective works, for resale or redistribution to servers or lists, or reuse of any copyrighted component of this work in other works.}
\newcommand\copyrightnotice{%
	\begin{tikzpicture}[remember picture,overlay]
		\node[anchor=south,yshift=10pt] at (current page.south) {\fbox{\parbox{\dimexpr\textwidth-\fboxsep-\fboxrule\relax}{\copyrighttext}}};
	\end{tikzpicture}%
}

\begin{document}

\title{Cable Optimization and Drag Estimation for Tether-Powered Multirotor UAVs}

\author{\IEEEauthorblockN{Max Beffert}
\IEEEauthorblockA{\textit{Computer Science Dept.} \\
\textit{University of Tübingen}\\
Tübingen, Germany \\
max.beffert@uni-tuebingen.de}
\and
\IEEEauthorblockN{Andreas Zell}
\IEEEauthorblockA{\textit{Computer Science Dept.} \\
\textit{University of Tübingen}\\
Tübingen, Germany \\
andreas.zell@uni-tuebingen.de}
}

\maketitle
\AddToHookNext{shipout/background}{\copyrightnotice}

\begin{abstract}
The flight time of multirotor unmanned aerial vehicles (UAVs) is typically constrained by their high power consumption. Tethered power systems present a viable solution to extend flight times while maintaining the advantages of multirotor UAVs, such as hover capability and agility. This paper addresses the critical aspect of cable selection for tether-powered multirotor UAVs, considering both hover and forward flight. Existing research often overlooks the trade-offs between cable mass, power losses, and system constraints. We propose a novel methodology to optimize cable selection, accounting for thrust requirements and power efficiency across various flight conditions. The approach combines physics-informed modeling with system identification to combine hover and forward flight dynamics, incorporating factors such as motor efficiency, tether resistance, and aerodynamic drag. This work provides an intuitive and practical framework for optimizing tethered UAV designs, ensuring efficient power transmission and flight performance. Thus allowing for better, safer, and more efficient tethered drones.
\end{abstract}


\section{Introduction}
One of the significant downsides of multirotor UAVs compared to fixed-wing aircraft is their high power demands, which often limit their flight time to less than 30 minutes. Nevertheless, they are helpful in many applications because of their ability to hover in place and their high agility. One solution to improve the flight time and retain some of the advantages of a multirotor is a tethered power system that supplies the UAV with power from the ground.\\
In prior work focusing on agricultural safety, we built a system in which a UAV is used to survey the surroundings of a ground vehicle for pedestrians \cite{beert_development_2024}. By using a tether-powered quadrocopter to hover over and follow an agricultural ground vehicle autonomously, we retain the advantages of a multirotor while allowing for indefinite flight times. Selecting the right cable is a significant aspect of building a tethered power system because it determines the limited power availability due to the tether resistance and the power requirements to lift the extra cable mass.\\
Although existing research on tether-powered UAVs is abundant, most studies do not go into detail regarding cable selection \cite{marques_tethered_2023}. A common approach is to use a high voltage on the tether to minimize losses. This approach is valid in many cases; however, it does not maximize efficiency and does not allow for optimization in extreme cases like using very long tethers. An additional benefit of selecting an appropriate tether is that it allows using lower voltages and improving safety.
Therefore, we propose an intuitive method for selecting optimum cables that can be applied to most tethered multirotor UAVs. Additionally, we propose a way to optimize not only for hover but also for forward flight and strong wind conditions by modeling the aerodynamic drag through system identification.

\section{Related Works}
\subsection{Cable Optimization}
In \cite{vishnevskiy_optimal_2017}, the number of conductors in a custom cable is optimized for use with high-voltage AC. The focus is on finding an efficient cable in general, not considering how the extra weight will affect the power requirements to lift the cable. This paper only applies to high-voltage AC, where finding a cable with a suitable impedance is a significant factor. In \cite{vishnevsky_experience_2019}, a similar approach is used to transmit very high power over a custom high voltage, high-frequency AC tether.\\
\cite{chang_design_2021} aims to calculate the maximum achievable flight height for tethers of different sizes. The downside of this approach is that there is usually a target flight height in practice, but some headroom is necessary to allow for agile maneuvers. By optimizing for flight height, it is not intuitive to see how big the margins are or how efficient the design is. In \cite{wang_optimum_2015}, the authors optimize cables of different lengths for carrying capacity. They talk about an experimental setup to determine the motor parameters but do not go into detail about the cable model that was used. The downsides of this approach are similar to \cite{chang_design_2021} since, in practice, there is usually a desired payload, and determining the necessary headroom is not intuitive.\\

\subsection{Drag Modeling}
Many modeling approaches for quadrotors exist, such as through system identification \cite{ivler_system_2019, ivler_multirotor_2019}, by blade element momentum (BEM) theory \cite{davoudi_hybrid_2019, park_dynamics_2024, bauersfeld_neurobem_2021} or using learning-based approaches \cite{mohajerin_deep_2018, akbari_computationally_2024}. For drag estimation, approaches intended for wind measurement are especially interesting since they require accurate drag modeling \cite{tagliabue_touch_2020, palomaki_wind_2017, meier_wind_2022, abichandani_wind_2020, mendez_wind_2023, noauthor_windspeed_nodate}.

\section{System Overview}
Tether-powered UAVs can be split roughly into two groups: those powered directly through the tether and those that use a high voltage on the tether, which is then converted to a lower voltage on the drone through a step-down converter. The direct approach has the benefit of low complexity, but the power that can be transmitted this way is limited by the maximum drone voltage. Therefore, this setup can only be used for short tethers. In this case, it is especially important to select the best possible cable. Looking at the power that can be transmitted through a given cable, we find that the formula for the direct and step-down approaches are identical except for a small loss factor caused by the voltage conversion. So, when selecting the optimum cable, both cases can be treated as one.\\
We consider flight heights of 30m, so powering the drone directly is not an option. Instead, we use a tether voltage of 75V DC that is stepped down to 16V on the drone and buffered with a battery. Having a battery is not strictly necessary but allows for emergency landings in case of tether failure and for more aggressive cable selection as it allows buffering short power spikes. This means that the tether does not need to be able to supply the maximum power spike requirements of the drone.\\
If not otherwise stated, all values are taken from our system, which is based on a Holybro X500 V2 quadrotor frame kit with 2216 KV920 motors and 10'' propellers. For the step-down converter, we use an off-the-shelf module with an assumed efficiency of 90\%. We use a Pixhawk 6c running ArduCopter firmware as a flight controller, but the proposed solutions should work on most other platforms. The complete system has a mass of 1.71kg in the tethered configuration (without the cable mass) and 1.55kg in the standalone configuration.

\begin{figure}[htbp]
	\centering
	\includegraphics[width=0.24\textwidth]{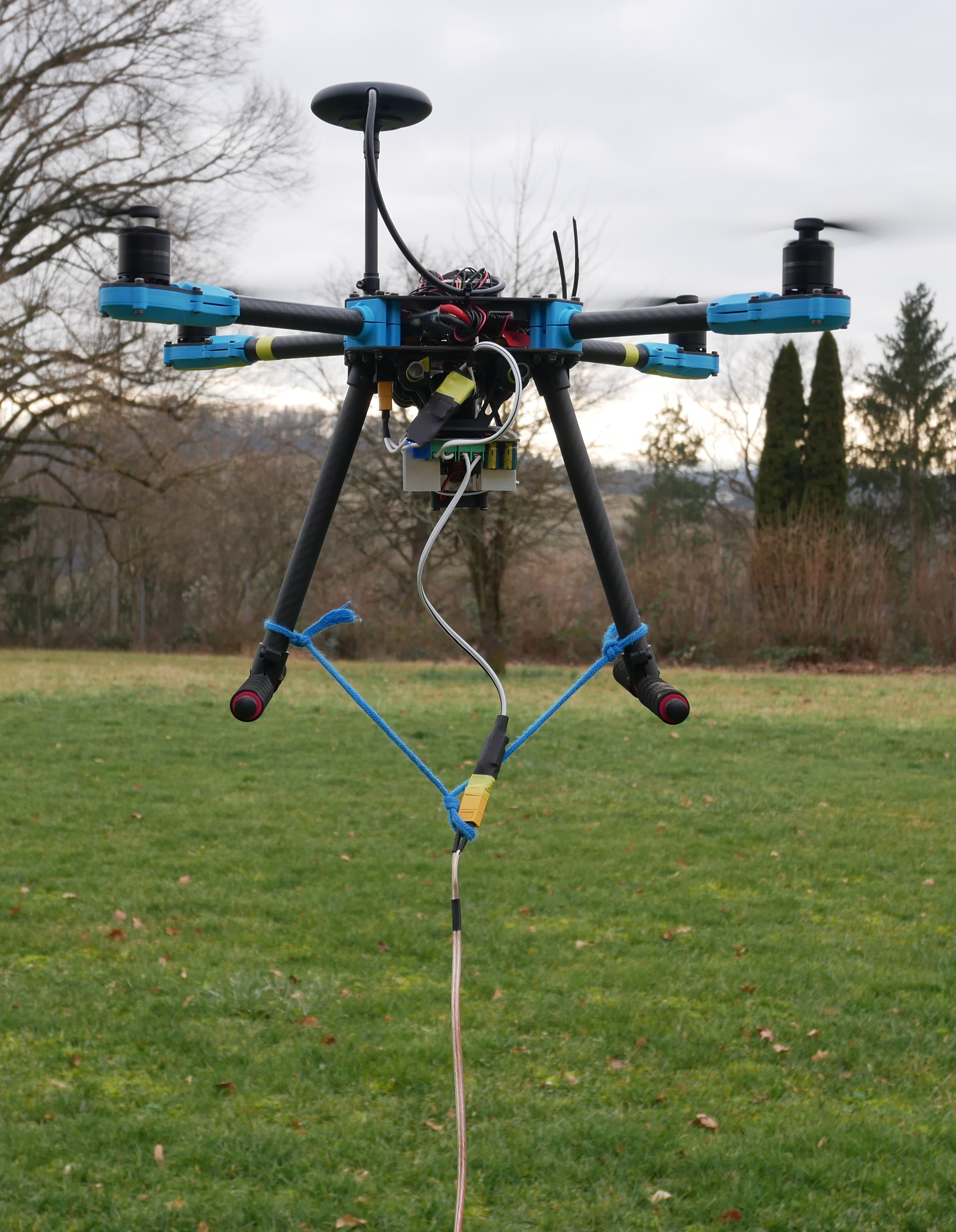}
	\caption{Finished system in flight}
\end{figure}

\section{Methods}
\subsection{Cable Optimization}
When dimensioning multirotor thrust requirements, a common rule of thumb is to ensure hover at half throttle. In a use case where more agility is necessary, this could be lower, but it generally should not be much higher.
Since motor efficiency decreases at higher throttle, a more general metric would be to ensure hover at half the maximum thrust. This approach can be directly applied to tethered multirotors by choosing a cable mass to have a thrust headroom of factor two. The downside of this approach is that it does not consider how much power can be transmitted over the cable. Depending on the drone configuration, this could lead to an oversized or undersized cable.\\
A more specific approach to tethered drones would be to dimension the cable to provide enough power for the motors at full throttle. This ensures power requirements but requires checking if the maximum thrust is enough to carry a cable of the calculated size. While this approach can not guarantee an efficient design, it is nonetheless helpful if a tethered drone without a buffer battery is constructed.\\
A hybrid between the two previous solutions is proposed to place a lower bound on the cable cross-section while maintaining the thrust requirements. This is achieved by calculating the maximum transmitted power and the available thrust achievable with that power depending on the cable cross-section. Additionally, the required thrust to lift the cable is calculated, making it possible to determine where the double thrust headroom is met while considering the transmitted power.\\
The minimum total power can be calculated and used as the optimal solution if it lies above the lower bounds of the thrust requirements. If a buffer is used, it might make sense to use the minimum total power even when the thrust headroom is not reached, as it is only fully used in highly dynamic maneuvers.\\
Knowing the motor efficiency curve is essential for cable optimization as it provides the relationship between motor efficiency and thrust. We used the curve given by the manufacturer, which often does not include data for throttle values below 50\%. In this case, we assume that the same efficiency applies to lower throttle values as well. The expectation is that the efficiency would be higher at a lower throttle (to a certain point), making this a conservative assumption.\\
The available thrust $T_{available}$ in kg per motor is given by:
\begin{equation}
T_{available}=motor(\eta_{DC}*(P_{available}-P_{add})/n)
\end{equation}
\begin{itemize}
	\item $motor(\cdot)$ is the motor curve, which maps the input power to the achieved thrust
	\item $\eta_{DC}$ is the efficiency of the DC-DC converter
	\item $P_{available}$ is the maximum available power
	\item $P_{add}$ is the additional power used by the flight controller and payload
	\item $n$ is the number of motors
\end{itemize}

\noindent As shown in \cite{chang_design_2021} and \ref{appendix}, the available power is limited by the tether resistance and voltage and can be calculated as follows:
\begin{equation}
P_{available}=U_{min}*(U-U_{min})/R_t\label{available_power_equation}
\end{equation}
\begin{itemize}
	\item $U_{min}=\max\{U_{minDC},\, U/2\}$ is the minimum received voltage based on the DC-DC input voltage cutoff
	\item $U$ is the tether input voltage
	\item $R_t$ is the tether resistance
\end{itemize}

\begin{equation}
R_t=2*h*\rho_{Rt}/A_t
\end{equation}
\begin{itemize}
	\item $h$ is the tether length and flight height
	\item $\rho_{Rt}$ is the conductor specific resistance
	\item $A_t$ is the conductor cross-section
\end{itemize}

\noindent The necessary thrust $T_{necessary}$ to lift the drone and cable in kg per motor is given by:
\begin{equation}
T_{necessary}=\cfrac{m_d+m_t}{\cos(\theta)*n}\label{necessary_thrust_equation}
\end{equation}
\begin{itemize}
	\item $m_d$ is the drone mass
	\item $m_t$ is the tether mass
	\item $\theta$ is the drone pitch angle (roll is assumed to be zero)
\end{itemize}

\begin{equation}
m_t=2*h*(\rho_{mt}*A_t+\rho_{mi}*A_i)
\end{equation}
\begin{itemize}
	\item $\rho_{mt}$ is the conductor density
	\item $\rho_{mi}$ is the insulation density
	\item $A_i$ is the insulation cross-section (without conductor)
\end{itemize}

\noindent The total power $P_{total}$ consisting of the power used by the drone $P_{UAV}=motor^{-1}(T_{necessary})$ and the power loss over the tether $P_{loss}$ is calculated using the tether current $I$ as follows:
\begin{equation}
P_{total}=U*I=P_{loss}+P_{UAV}=I^2*R_t+P_{UAV}
\end{equation}
Which results in the quadratic equation:
\begin{equation}
I^2*R_t-U*I+P_{UAV}=0
\end{equation}
That can be solved as follows (ignoring negative result):
\begin{equation}
P_{total}=U*\cfrac{U-\sqrt{U^2-4R_t*P_{UAV}}}{2R_t}
\end{equation}

\subsection{Drag Estimation}
The previously described cable optimization is easy for a scenario where the drone hovers in place but gets more complicated if done for a drone in forward flight or in case of strong wind, as this requires estimating the drag at the target airspeed.\\
Multiple methods are commonly used to estimate the drag of multirotor UAVs. The most straightforward and performant is to measure the frontal area and assume a coefficient of friction to calculate the drag of the drone at different speeds directly. One issue with this approach is that it does not include the induced drag from the propellers and requires assumptions about the drag coefficient.\\
The blade element momentum theory is another common approach that has shown excellent accuracy since it includes the aerodynamic properties of the propellers. The downside of this approach is that it requires exact measurement and knowledge of the used propellers.\\
Wind tunnel measurements are another option to get precise data, but they require a lot of effort and expense. A cheaper alternative that can provide similar results is using computational fluid dynamics simulations. However, this requires making an accurate drone model, which is hard to verify, especially regarding dynamics.\\
We propose using a system identification approach. System identification on drones can be complex if the focus is completely modeling the drone's behavior. In this case, the model can be much simpler since we are only interested in modeling the drag in static forward flight. Our approach is similar to \cite{meier_wind_2022}, which uses known angle-dependant drag coefficients in combination with a linear or quadratic velocity term to estimate the wind effects. Instead of using known drag coefficients and estimating the wind, we want to estimate the drag coefficients for known wind conditions. Since we do not measure the airspeed, flying in low wind and consistent conditions is crucial, as well as using a symmetric testing approach and a symmetric model to ensure the wind effects can be estimated.\\
The drag force $F_D$ can be calculated as the difference between the observed forward acceleration $a_{of}$ and expected forward acceleration $a_{ef}$, multiplied by the drone mass:
\begin{align}
a_{ef}&=(g+a_{od})*\tan(\theta)/cos(\Phi)\\
F_D&=m*(a_{of}-a_{ef})
\end{align}
\begin{itemize}
	\item $g$ is the gravitational constant
	\item $a_{od}$ is the observed acceleration in downward direction
	\item $\theta$ is the pitch angle and $\Phi$ is the roll angle
	\item $m$ is the mass of the drone
\end{itemize}

\begin{figure}[htbp]
	\centering
	\begin{subfigure}[t]{0.3\textwidth}
		\centering
		\caption{dynamic}
		\vspace*{5pt}
		\includegraphics{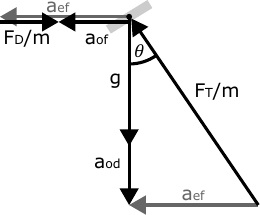}
		\label{dynamic}
	\end{subfigure}%
	\begin{subfigure}[t]{0.2\textwidth}
		\centering
		\caption{static}
		\vspace*{5pt}
		\includegraphics{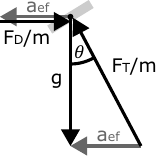}
		\label{static}
	\end{subfigure}
	\caption{Acceleration balance during forward flight}
	\label{forces}
\end{figure}

We have to assume that there is no vertical wind and drag, so the vertical velocity should be kept to a minimum during testing. It would be possible to consider these factors, but this would require more training data and might reduce the model accuracy for forward drag \cite{meier_wind_2022}.\\
Our drag model is based on the classic drag equation \cite{isele_formeln_2017}
\begin{equation}
F_D=1/2*\rho_{air}*C_D*A*v^2
\end{equation}

\noindent The goal is to estimate the drag force from the drone's pitch angle and airspeed. 
The pitch angle depends on the drone's airspeed, so estimating the drag from the airspeed alone would theoretically be possible. However, this would limit the training data to using only time slices with zero acceleration. Another downside is that it would not generalize well if the drone mass is changed since that also influences the angle at a given airspeed.
The part of the equation dependent on the pitch angle $\theta$ is the projected area $A$, which is modeled as an effective area calculated from the front area $A_f$ and top area $A_t$, which are learned model parameters. These also include the drag coefficient, so it cannot be assumed to be the actual area of the physical drone.
\begin{equation}
A(\theta)=A_f*\cos{|\theta|}+A_t*\sin{|\theta|}
\end{equation}

\begin{figure*}[ht]
	\centering
	\begin{subfigure}[b]{0.4\textwidth}
		\centering
		\caption{without linear velocity term}
		\includegraphics[width=\textwidth]{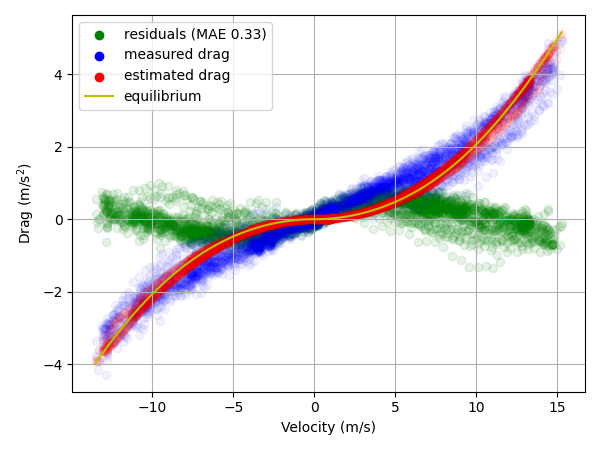}
		\label{fig:nolin}
	\end{subfigure}%
	\begin{subfigure}[b]{0.4\textwidth}
		\centering
		\caption{with linear velocity term}
		\includegraphics[width=\textwidth]{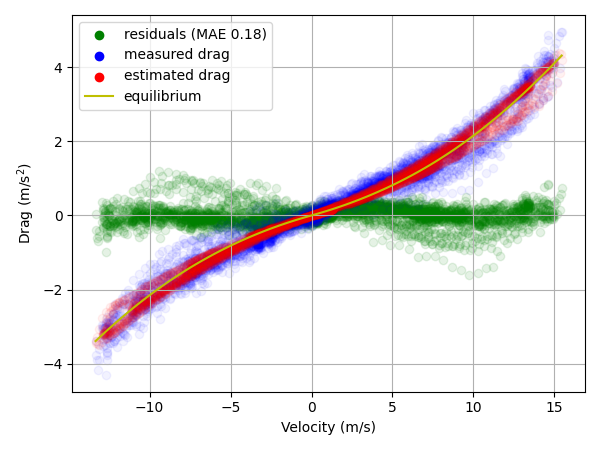}
		\label{fig:lin}
	\end{subfigure}
	\caption{Flattened drag estimation as a function of velocity. The blue line represents the measured drag, red shows the estimated drag, and green represents the residuals. The yellow line indicates the drag during static forward flight (no acceleration). Comparing (a) and (b) demonstrates that combining the quadratic drag equation with a linear term is necessary for accurate modeling.}
	\label{2d_drag}
\end{figure*}

\noindent After implementing this solution, it became apparent that the quadratic model was insufficient to describe the drag correctly (see Fig. \ref{2d_drag}). For this reason, an additional linear term with learned parameter $j$ was added.
This observation is consistent with the findings from \cite{meier_wind_2022} that a linear model is more accurate for low airspeed and a quadratic for high airspeed. By combining both, we can accurately estimate the drag over a wide airspeed range.
Additionally, a learned velocity offset $w$ was used to model wind effects and estimate airspeed from drone velocity.
\begin{equation}
F_D(v, \theta)=1/2*\rho_{air}*A(\theta)*((v-w)^2+j(v-w))
\end{equation}
\begin{itemize}
	\item $\rho_{air}$ is the air density
	\item $v$ is the forward velocity
\end{itemize}

\noindent The resulting function describes the drag at different airspeeds and pitch angles and can be used to model a dynamic flight (with acceleration). In static forward flight, there is a force equilibrium where the vertical component of the thrust has to counteract gravity, and the horizontal component counteracts the drag. This lets us calculate the pitch angle during forward flight by numerically solving the equation for a given target velocity.
\begin{equation}
	F_D(v, \theta)+F_{Dt}(v)=(m_d+m_t)*g*\tan{\theta}
\end{equation}
\begin{itemize}
	\item $F_{Dt}$ is the tether drag
	\item $m_d$ is the drone mass
	\item $m_t$ is the tether mass
\end{itemize}

\noindent Our implementation of the cable optimization and drag estimation is available on GitHub \cite{beffert_cable_nodate}.

\section{Experiments}
Generally, it is advisable to use aluminum cables as they have a better ratio of resistance to weight (Table \ref{materialComparison}), allowing for cables with about half the conductor weight compared to copper. We assume the insulation is made from PVC and has a constant thickness of 0.75mm. In reality, the insulation thickness does not seem to be based on the conductor diameter, but this relationship could be considered.

\begin{table}[bp]
	\centering
	\caption{Specific resistance, density and combined material constant for different conductor materials \cite{isele_formeln_2017}}
	\begin{tabular}{ c | c c c}
		&$\rho_r$&$\rho_m$ & $\rho=\rho_r*\rho_m$\\
		&$[\frac{\Omega mm^2}{m}]$&$[\frac{g}{cm^3}]$&$[\frac{\Omega g}{m^2}]$\\[5pt]
		\hline\\[-5pt]
		Copper  & 0.0178 & 8.9 & 0.158 \\
		Aluminum& 0.0278 & 2.7 & 0.075\\
	\end{tabular}
	
	\label{materialComparison}
\end{table}

\noindent The required data for drag estimation is collected from two flights in the standalone configuration (without the tether).\\
In the first flight, the drone is hovered in place at different yaw angles during low wind conditions. By doing this, an offset can be calculated for the pitch angle by averaging the pitch over the entire flight duration. This step is optional and can be skipped if the gyroscope is well-calibrated.\\
During the second flight, the drone is accelerated forward and backward at different pitch angles. By going in opposite directions, most wind influence can be estimated in the final model as long as it is constant. A flight mode with altitude hold is advised for testing to reduce vertical aerodynamic effects. Flight modes that allow control over the angle are preferred since they provide sparse angle data but dense velocity data. This is beneficial since the effect of the velocity on drag is much bigger than that of the pitch angle. GPS-based stabilized flight modes can also be used but are less desirable as they provide dense angle data and sparse velocity data.\\
Data from the ArduCopter black box log is used, specifically the pitch, yaw, and velocity output from the Extended Kalman Filter (EKF). The data does not undergo additional smoothing or filtering. This approach makes it easy to adapt to any drone configuration and flight controller software. It would be possible to gather this data wirelessly through telemetry and run the model during flight. The only additional knowledge that is required about the drone is its mass.\\
The velocity vector is in world frame, so it must first be converted to a stabilized drone frame. This means that the frame is aligned to the yaw of the drone but not the pitch and roll (the forward and left axes are parallel to the ground). Then, differentiating the velocity gives the drone's observed acceleration.\\

\begin{figure}[htbp]
	\centering
	\includegraphics[width=0.5\textwidth]{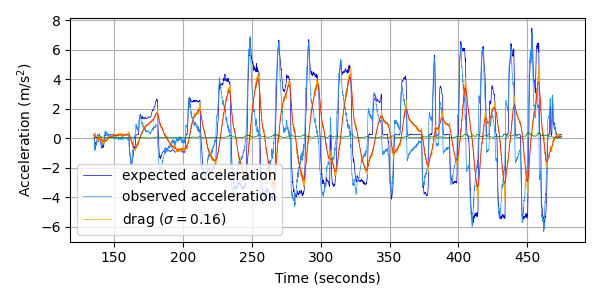}
	\caption{Input data from a single flight. Dark blue is the expected acceleration calculated from the drone pitch. Light blue is the observed acceleration from the onboard accelerometer and GPS sensors (EKF). Orange is the drag calculated as the difference between expected and observed acceleration.}
\end{figure}

\section{Results}
\subsection{Cable Optimization}
The point where the thrust requirement is reached [$T_{available}(A_t)=2*T_{necessary}(A_t)$] and the point of minimum power [$\min P_{total}(A_t)$] are determined numerically.
The one with the bigger conductor cross-section is selected as the final optimal solution.\\

\begin{figure}[tbp]
	\centering
	\includegraphics[width=0.5\textwidth]{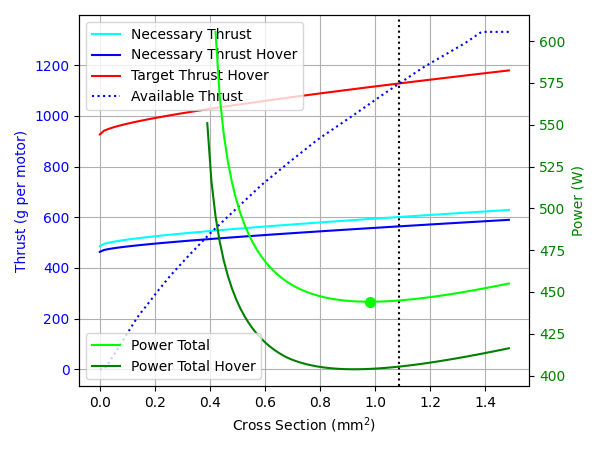}
	\caption{Available vs necessary thrust for hover at 30m and forward flight at 30 km/h using Holybro 2210 motors with 10'' propellers, as well as the required power. The green dot marks the minimum power needed for forward flight, while the black dotted line indicates the required thrust headroom. In this case, a $1.09mm^2$ cable is optimal.}
	\label{holybro}
\end{figure}

\begin{figure}[htbp]
	\centering
	\includegraphics[width=0.5\textwidth]{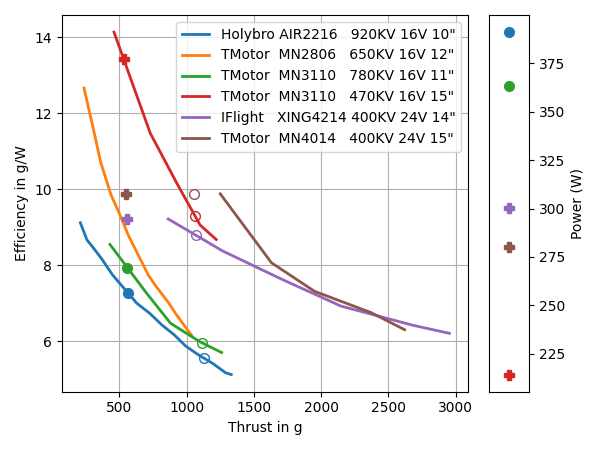}
	\caption{Comparison between different motor models and their optimum tethers. Circle is required thrust at hover, dot is thrust headroom (two times hover thrust), + is where minimum power is optimal (30km/h forward flight)}
	\label{comparison}
\end{figure}

Besides selecting the optimum cable, the model also allows for comparing different motor configurations  (Fig. \ref{comparison}). In this case, we need to assume a known drone frame. In reality, this is not perfectly accurate, as bigger rotors need bigger and potentially heavier frames. Nevertheless, this can be useful to find unfeasible combinations.\\
Interestingly, more efficient motors are not necessarily better if they cannot produce enough thrust to satisfy the constraints. So, the perfect motor depends mainly on the frame and payload weight, but a high efficiency is desirable. Therefore, multirotors with bigger propellers, which are generally more efficient, are preferable for use with a tethered power system.

\subsection{Drag Estimation}

The parameters of the drag model $A_f$, $A_t$, $j$ and $w$ are estimated using nonlinear least squares:
\begin{align*}
	A_f&=1.492*10^{-2} m^2 \hspace{-50pt}&\pm3.07\%\\
	A_t&=1.368*10^{-2} m^2 \hspace{-50pt}&\pm3.18\%\\
	j&=19.43 m/s \hspace{-50pt}&\pm4.35\%\\
	w&=-1.196 m/s \hspace{-50pt}&\pm1.61\%
\end{align*}

\begin{figure}[htbp]
	\centering
	\includegraphics[width=0.5\textwidth]{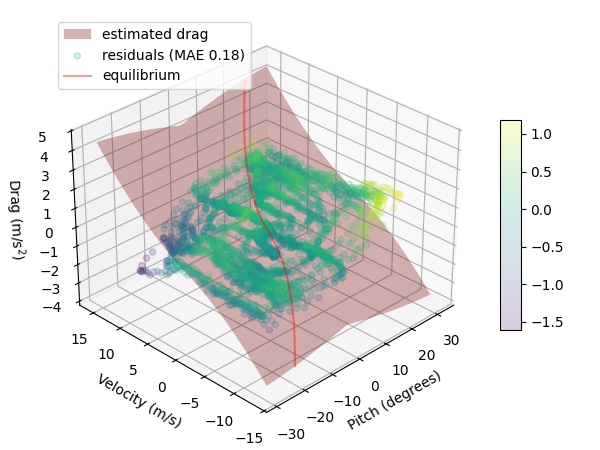}
	\caption{The red plane represents the learned drag function based on pitch and forward velocity. The red line shows the drag during static forward flight (no acceleration). The points indicate the residuals of the measured drag. Residuals are larger when the drone flies in one direction while pointing in the opposite direction to decelerate due to turbulent airflow from the propellers (propwash).}
	\label{3d_drag}
\end{figure}

\noindent It is important to note that the wind velocity offset is specific to each flight. Hence, it must be estimated separately when using data from multiple flights. The wind offset is set to zero for the final model to get a result without any wind.\\
The estimated drag function can be best understood by looking at the 3D plot (Fig. \ref{3d_drag}). Here, we see a uniform residual distribution, which mostly results from the noisy acceleration data used in the input. The only exception is higher residuals in the quadrants where the velocity sign is opposite that of the pitch. Here, the drone is flying in one direction but is already pointing in the other direction to decelerate. This causes the drone to fly into the turbulent airflow caused by the propellers. This phenomenon is called propwash, and modeling it would require adding additional asymmetric terms to the system model. In this case, we are only interested in the drag at equilibrium, so any inaccuracies due to propwash can be disregarded. \\
The drag of the tether is modeled through the drag equation as two independent cylinders. The actual drag coefficient of two connected cylinders is probably slightly higher. The area is multiplied by 0.75 to account for the random rotation of the two tether conductors. It is assumed that all the tether drag force acts on the drone. This happens when the drone flies far in front of the ground vehicle, representing the worst case. For a more precise model, the relative position of the drone and the ground vehicle and the resulting tether shape could be used to calculate the exact effect of the tether drag force on the drone.\\
By modeling the drag forces in forward flight, we can calculate the resulting pitch angle in equilibrium for (\ref{necessary_thrust_equation}). This results in a bigger required thrust than during hover. When selecting the optimal cable, this mainly applies to the point of least power, allowing the system to be optimized for power consumption in forward flight. It could also be used when calculating the double thrust headroom, but that rule of thumb is usually used in relation to the hover thrust.\\
Applying the approach to our specific UAV, we determined that using an aluminum cable with a cross-section of 1.09$mm^2$ is optimal for a tether voltage of 75V (Fig. \ref{holybro}).
This results in a tether mass of 0.55kg for a combined system mass of 2.26kg with 405W power draw in hover and 445W at 30km/h flight speed.\\

\newpage
\section{Conclusion}
This work presents a comprehensive approach to optimizing cable selection for tether-powered multirotor UAVs, considering both hover and forward flight conditions. Our method improves on previous work by optimizing for power usage as well as system constraints while proving an intuitive metric to gauge how the selected cable will affect the system performance.\\
Through experiments, we tested our approach by analyzing various motor configurations and their interaction with different cable sizes. The results highlight the importance of selecting an appropriate motor and tether combination to maximize performance while minimizing energy losses.\\
Using angle-based flight modes instead of speed-based ones, we obtained more dense velocity data and faster acceleration, reducing the required space and time for the experiments.\\
Additionally, we improved on previous work by incorporating forward flight drag forces, allowing for more targeted optimization of tethered UAV systems. Using physics-informed modeling with system identification techniques, we developed a method to effectively estimate drag from a single flight under real-world conditions, only requiring the drone mass as additional knowledge and no onboard or external aerodynamic sensors.\\
One limitation of this work is that many assumptions have been made about the tether's aerodynamics and how it affects the drone. Future work could improve this by estimating the tether drag coefficients alongside the drone's aerodynamics. Another avenue of improvement would be to consider the drone's relative position to the ground vehicle and how the resulting tether shape affects the forces applied to the drone. This would enable real-time estimations and model predictive control of the tethered UAV.



\appendix{Derivation of Equation \ref{available_power_equation}:}
\label{appendix}
The voltage drop over the tether $U_t$ can be calculated from the tether input voltage $U$ and the voltage drop over the drone $U_{UAV}$.\\
\begin{equation}
\begin{aligned}
&U_t=U-U_{UAV}\\
&I=I_{UAV}=I_t=U_t/R_t\\
&P_{UAV}=U_{UAV}*I=U_{UAV}*(U-U_{UAV})/R_t\\
\end{aligned}
\end{equation}
Assuming the maximum available power occurs at the minimum DC-DC converter input voltage ($U_{UAV}=U_{minDC}$), we obtain the equation presented in \cite{chang_design_2021}, which is valid for $U_t\leq U_{minDC}$. To show this, we find the point where $P_{UAV}$ is maximum. This occurs where the first derivative with respect to $U_{UAV}$ is zero:\\
\begin{equation}
	\begin{aligned}
		&\cfrac{\mathrm{d}P_{UAV}}{\mathrm{d}U_{UAV}}=\cfrac{U-2U_{UAV}}{R_t}=0\\
		&U_{UAV}=U/2=U_t \quad(\text{given}\; R_t>0)
	\end{aligned}
\end{equation}
This means that the maximum power transfer occurs when the voltage drop over the UAV equals the voltage drop over the tether.
Since the minimum voltage at the UAV is limited by the DC-DC converter input, we say that:
\begin{equation}
\begin{aligned}
&U_{min}=\max\{U_{minDC},\, U/2\}\\
&P_{available}=U_{min}*I=U_{min}*(U-U_{min})/R_t
\end{aligned}
\end{equation}
With $P_{available}$ being the maximum power that can be used by the drone.

\bibliographystyle{IEEEtran}
\bibliography{references}

\end{document}